\documentclass[letterpaper]{article} % DO NOT CHANGE THIS
\usepackage{aaai25}  % DO NOT CHANGE THIS
\usepackage{times}  % DO NOT CHANGE THIS
\usepackage{helvet}  % DO NOT CHANGE THIS
\usepackage{courier}  % DO NOT CHANGE THIS
\usepackage[hyphens]{url}  % DO NOT CHANGE THIS
\usepackage{graphicx} % DO NOT CHANGE THIS
\usepackage{natbib}  % DO NOT CHANGE THIS AND DO NOT ADD ANY OPTIONS TO IT
\usepackage{caption} % DO NOT CHANGE THIS AND DO NOT ADD ANY OPTIONS TO IT
\usepackage{algorithm}
\usepackage{algorithmic}
\usepackage{fontawesome}
\usepackage{multirow}
\usepackage[table,xcdraw]{xcolor}
\usepackage{array,multirow,graphicx}
\usepackage{multicol}
\usepackage{multirow}
\usepackage{makecell}
\usepackage{booktabs}
\usepackage{fixltx2e}
\usepackage{tabularx}
\usepackage{calc}
\usepackage{colortbl}
\usepackage{xcolor}
\usepackage{newfloat}
\usepackage{listings}
\DeclareCaptionStyle{ruled}{labelfont=normalfont,labelsep=colon,strut=off} % DO NOT CHANGE THIS
\floatstyle{ruled}
\newfloat{listing}{tb}{lst}{}
\floatname{listing}{Listing}
\title{\faBolt \hspace{0.1cm} SPARK: Multi-Vision Sensor Perception and Reasoning Benchmark \\ for Large-scale Vision-Language Models}
\author{
Youngjoon Yu$^\dagger$, Sangyun Chung$^\dagger$, Byung-Kwan Lee, and Yong Man Ro\thanks{Corresponding author. $^\dagger$ \textnormal{Both authors are equally contributed.}}
}

\affiliations {
    Integrated Vision Language Lab, KAIST, South Korea \\
    \{greatday, jelarum, leebk, ymro\}@kaist.ac.kr
}

\begin{document}

\maketitle

\begin{abstract}
Large-scale Vision-Language Models (LVLMs) have significantly advanced with text-aligned vision inputs. They have made remarkable progress in computer vision tasks by aligning text modality with vision inputs. There are also endeavors to incorporate multi-vision sensors beyond RGB, including thermal, depth, and medical X-ray images. However, we observe that current LVLMs view images taken from multi-vision sensors as if they were in the same RGB domain without considering the physical characteristics of multi-vision sensors. They fail to convey the fundamental multi-vision sensor information from the dataset and the corresponding contextual knowledge properly. Consequently, alignment between the information from the actual physical environment and the text is not achieved correctly, making it difficult to answer complex sensor-related questions that consider the physical environment. In this paper, we aim to establish a multi-vision \textit{\textbf{S}}ensor \textit{\textbf{P}}erception \textit{\textbf{A}}nd \textit{\textbf{R}}easoning benchmar\textit{\textbf{K}} called \textit{\textbf{SPARK}} that can reduce the fundamental multi-vision sensor information gap between images and multi-vision sensors. We generated 6,248 vision-language test samples to investigate multi-vision sensory perception and multi-vision sensory reasoning on physical sensor knowledge proficiency across different formats, covering different types of sensor-related questions. We utilized these samples to assess ten leading LVLMs. The results showed that most models displayed deficiencies in multi-vision sensory reasoning to varying extents. Codes and data are available at \textbf{https://github.com/top-yun/SPARK}
\end{abstract}

\section{Introduction}
In recent days, Large-scale Vision-Language Models (LVLMs) have achieved significant breakthroughs in areas such as visual dialogue~\cite{koh2023grounding}, video analysis~\cite{ren2024timechat}, and document understanding~\cite{ye2023mplug}, establishing themselves as critical tools in the pursuit of artificial general intelligence (AGI). These models function similarly to the human brain by processing multimodal information and generating sophisticated inferences. For instance, the latest LVLMs, like OpenAI's GPT-4o~\cite{gpt4oblog}, have exhibited exceptional reasoning abilities that not only rival but in some cases exceed human performance.

%%%%%%%%%%%%%%%%%%%%%%%%%%%%%%%%%%%%%%%%%%%%%%%%%%%%%%%%%%%%
\begin{figure}[t!]
  \centering
  \includegraphics[width=1.0\linewidth]{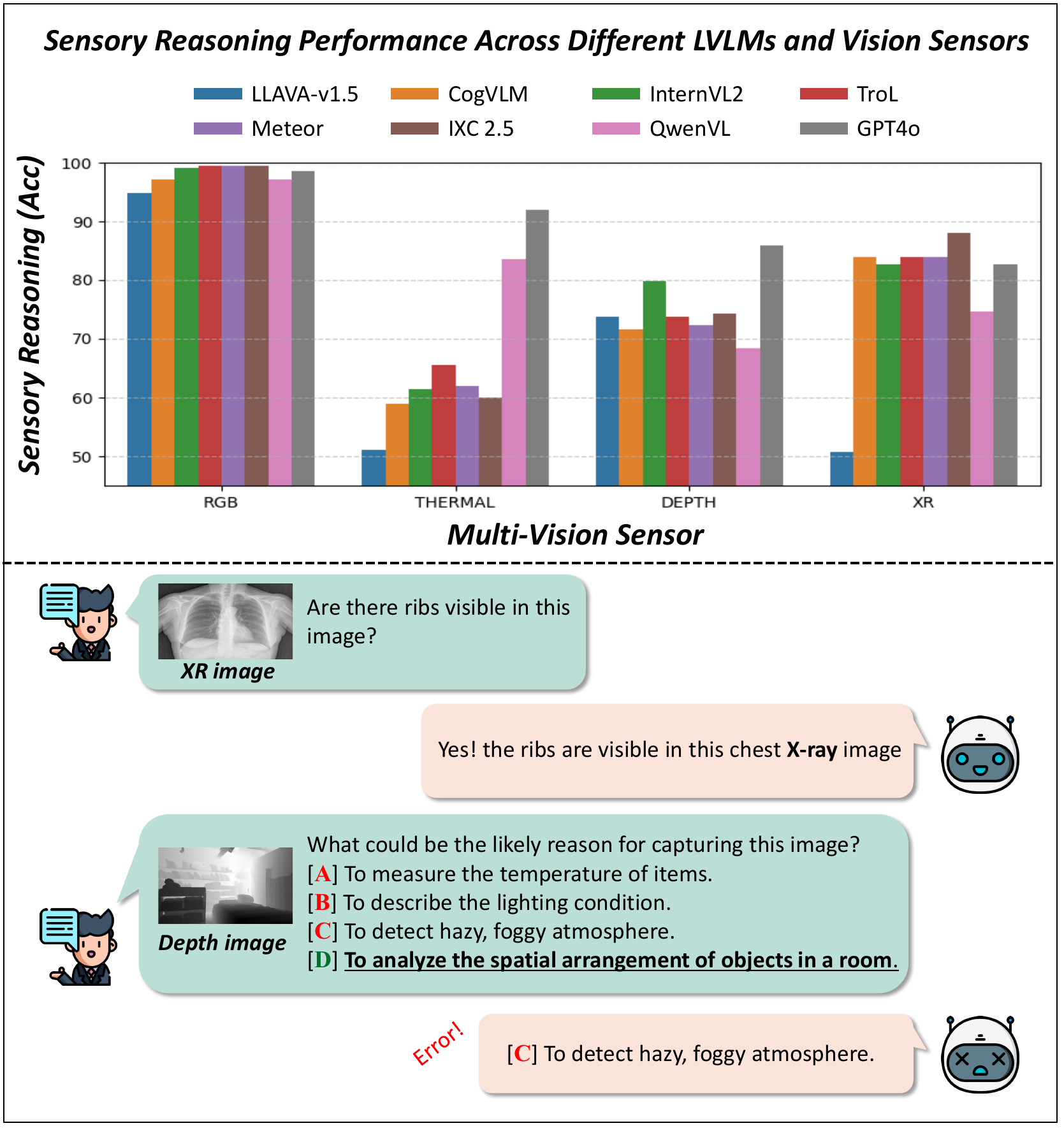}
  \caption{The comparison of sensory reasoning performance across different multi-vision sensors with respect to the recent LVLMs. Note that, sensory reasoning performance significantly drops across different multi-vision sensors.}
  
  \label{fig:1}
\end{figure}
%%%%%%%%%%%%%%%%%%%%%%%%%%%%%%%%%%%%%%%%%%%%%%%%%%%%%%%%%%%%

One emerging concept in modern AI research gaining significant attention is the development of large vision language models (LVLMs) capable of handling a variety of multimodal inputs, surpassing the capabilities of previous large language models (LLMs). LVLMs can process diverse forms of data simultaneously, including images, videos, and text ~\cite{gpt4oblog, internvl2blog, zhang2024internlmxcomposer25versatilelargevision}. This ability also allows them to use multi-vision sensor data as input, including thermal sensors, depth sensors, and medical imaging~\cite{girdhar2023imagebind, su2023pandagpt}. To fully harness the potential of LVLMs, recent research has focused on effectively integrating various multi-vision sensor data to develop more sophisticated and practical AI systems for the real world. 

However, despite the remarkable advancements in LVLM models, significant challenges still remain in fully utilizing multi-vision sensors. LVLMs often overlook the nuances of the physical properties of individual vision sensors. Instead, they tend to make judgments based on prior visual or linguistic information from images they have learned using low-level features in two-dimensional data. This results in the models recognizing only superficial patterns in image inputs, missing the underlying logical structures or contextual understanding. When identifying specific objects in an image input, a model might rely on patterns learned from similar-looking images rather than considering the actual physical properties of the multi-vision sensors used to capture the image. This can hinder accurate identification and a deep understanding of the input images in fields where the LVLM's decision-making is crucial such as autonomous driving~\cite{mao2023gpt,xu2024drivegpt4}, security systems~\cite{shi2024shield}, and medical image diagnosis~\cite{bazi2023vision}.

%%%%%%%%%%%%%%%%%%%%%%%%%%%%%%%%%%%%%%%%%%%%%%%%%%%%%%%%%%%%
\begin{figure*}[t!]
  \centering
  \includegraphics[width=1.0\linewidth]{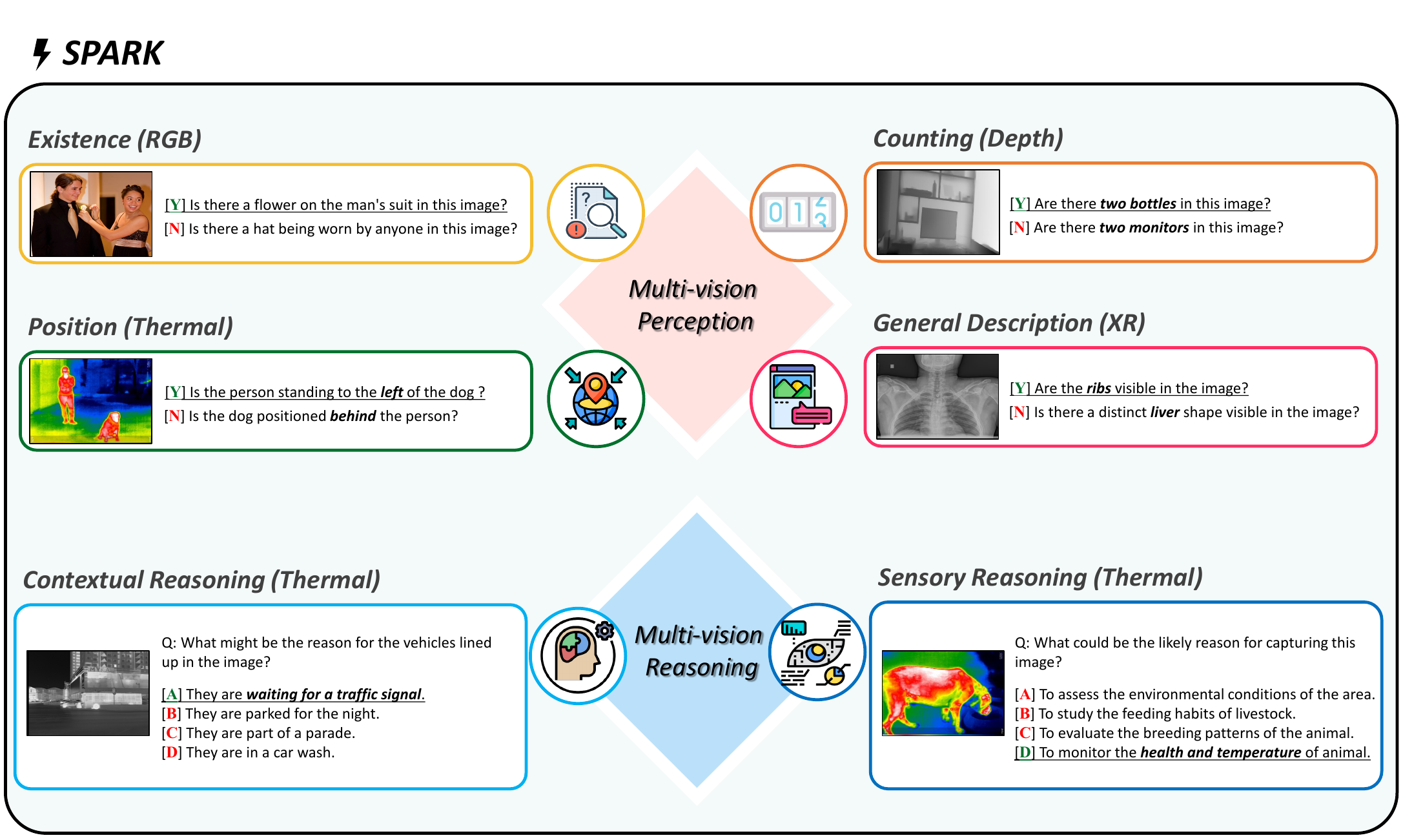}
  \caption{In the proposed SPARK, we build the first benchmark for evaluating the abilities of LVLMs in multi-vision sensor understanding, which covers four types of multi-vision perception tasks (Existence, Counting, Position, and General Description) and two types of multi-vision reasoning tasks (Contextual Reasoning and Sensory Reasoning).}
  \vspace{-0.4cm}
  \label{fig:overall}
\end{figure*}
%%%%%%%%%%%%%%%%%%%%%%%%%%%%%%%%%%%%%%%%%%%%%%%%%%%%%%%%%%%%

We evaluate the behavior of the recent LVLMs using multi-vision sensor images as input in Figure 1. The performance of sensory reasoning, which we devised to assess the understanding of fundamental knowledge of multi-vision sensors in the real world, significantly drops across different multi-vision sensors such as thermal infrared, depth, and X-ray (XR) images. This highlights the challenges that LVLMs face in accurately interpreting multi-vision sensor data and making correct inferences based on the physical properties of sensors. Additionally, from the interaction example shown below in Figure 1, while the LVLM can accurately identify the vision sensor used to capture the image for a relatively simple question, it struggles with understanding the actual purpose or context of the image in the sensor-related, more complicated questions. This indicates that current LVLMs have difficulty in understanding the fundamental knowledge of physical vision sensors beyond what the image looks like.

For example, as illustrated in Figure 1, when humans look at a photograph of an X-ray medical image, they interpret it deeply, drawing upon their knowledge base and their physical understanding of the human body beyond the X-ray image itself. Despite never having seen their internal organs and the structure of bones with the naked eye, humans can comprehend the image through scientific contextual knowledge and their inherent understanding of the physical world. In contrast, current LVLMs try to understand the inside of the human body based solely on the two-dimensional data they have been trained on, revealing their limitations in fully grasping the physical environment of the real world. Therefore, establishing a comprehensive evaluation benchmark is necessary before LVLMs are implemented in critical and sensitive real-world applications. However, the assessment of Large Vision-Language Models (LVLMs) has significantly lagged behind their rapid development. Several initiatives are striving to close this gap by introducing a variety of multimodal evaluation benchmarks. Notable examples include MME~\cite{fu2024mmecomprehensiveevaluationbenchmark}, MMBench~\cite{liu2024mmbenchmultimodalmodelallaround}, LVLM-eHub~\cite{xu2023lvlmehubcomprehensiveevaluationbenchmark}, and  SEED-Bench~\cite{li2023seed}. These benchmarks aim to define key dimensions of multimodal capabilities and provide corresponding test samples. But, they cover a relatively narrow range of multimodal tasks, primarily focusing on fundamental abilities such as visual recognition and OCR.

In this paper, to handle the aforementioned challenge, we design the SPARK benchmark to evaluate multi-vision input LVLMs on two fronts: multi-vision perception and multi-vision reasoning. Multi-vision perception pertains to the information needed, which measures the LVLM's effectiveness in satisfying visual perception needs. Multi-vision reasoning measures the LVLM's ability to base its responses on fundamental information from the provided sensor knowledge. To be specific, we generated 6,248 vision-language test samples to investigate multi-vision sensory perception and reasoning related to physical sensor knowledge proficiency, covering 6 types of multi-vision sensory instruction tasks across 2 different question-and-answer formats. We used these samples to assess 10 leading large-scale vision language models. The experiment results validate that most LVLMs displayed deficiencies in sensory reasoning to varying extents. 

In summary, the contributions of this work are as follows:
\begin{itemize}
\item To the best of our knowledge, we first reveal the incapability of current LVLMs, which suffer from limited multi-vision sensory reasoning across different multi-vision sensors due to an absence of fundamental understanding of sensors in the physical world.

\item We propose a novel benchmark, SPARK, to rigorously test and evaluate the capabilities of LVLMs in understanding sensory knowledge, providing a comprehensive framework for assessing their performance.

\item We evaluated a total of 10 state-of-the-art LVLMs using our SPARK benchmark, which is designed to rigorously assess the capability of the LVLMs in handling fundamental knowledge related to multi-vision sensors.

\end{itemize}

\section{Related work}
\noindent \textbf{Large-scale Vison-Language Models.} Recently, there has been significant interest in visual language multimodal learning. Visual language models such as LLAVA~\cite{liu2023llava, liu2024llavanext}, CollaVO ~\cite{lee2024collavocrayonlargelanguage}, MoAI~\cite{lee2024moaimixtureintelligencelarge}, TroL~\cite{lee2024troltraversallayerslarge}, Meteor~\cite{lee2024meteormambabasedtraversalrationale}, IXC2.5~\cite{zhang2024internlmxcomposer25versatilelargevision}, and QwenVL ~\cite{bai2023qwenvlversatilevisionlanguagemodel} have shown impressive performance in a variety of downstream tasks. 
In addition, to obtain richer contextual information, LVLMs have developed the capability to handle multimodal inputs. \citeauthor{wang2023cogvlm} introduces CogVLM, an advanced visual language foundation multimodal model that integrates a trainable visual expert module with a pretrained language model. InternVL2~\cite{chen2024far} is an open-source multimodal large language model that bridges the gap between open-source and commercial models by enhancing visual understanding, dynamic high-resolution processing, and bilingual dataset quality. GPT4o~\cite{gpt4oblog} possesses advanced multimodal capabilities, allowing it to process and generate diverse multimodalities. This enables the model to understand and create content that integrates visual and textual information, making it suitable for a wide range of applications that require various modalities. Consequently, many LVLMs have emerged that take multi-vision sensor images as input. \citeauthor{girdhar2023imagebind} presents ImageBind, which creates a joint embedding space across multi-vision sensors including depth and thermal sensor data. PandaGPT~\cite{su2023pandagpt} is a LVLM that integrates multimodal encoders and large language models to enable multi-vision and auditory instruction-following capabilities, performing complex tasks. However, relatively less attention has been devoted to whether LVLMs truly understand the physical meanings of multi-vision sensors used to capture the input image.

\noindent \textbf{Evaluation Benchmark for LVLMs.} Numerous studies have leveraged existing vision-language datasets to develop benchmarks for assessing the reliability of LVLMs~\cite{li2024survey}. MME ~\cite{fu2024mmecomprehensiveevaluationbenchmark} includes 14 sub-tasks based on publicly available images with manually created annotations, evaluating both the recognition and perception capabilities of LVLMs through yes/no question answering. SEED-benchmark~\cite{li2023seed} designed to evaluate the generative comprehension capabilities of multimodal LVLM through human-annotated multi-choice questions across 12 evaluation dimensions. Other comparable benchmarks include LVLM-eHub~\cite{xu2023lvlmehubcomprehensiveevaluationbenchmark}, MM-Vet~\cite{yu2023mmvetevaluatinglargemultimodal}, and MMBench~\cite{liu2024mmbenchmultimodalmodelallaround}. Additionally, there are benchmarks aimed at assessing specific target properties of LVLMs. POPE~\cite{li2023POPE} focuses on evaluating object hallucination by asking yes/no questions about the presence of objects in the input image. M-HalDetect~\cite{gunjal2024detectingpreventinghallucinationslarge} introduces hallucination tasks using human-annotated labels for sentence-level classification. Unlike those previous evaluation benchmarks, the proposed SPARK is designed to rigorously test and evaluate the capabilities of understanding the physical meaning of multi-vision sensors.

%%%%%%%%%%%%%%%%%%%%%%%%%%%%%%%%%%%%%%%%%%%%%%%%%%%%%%%%%%%%
\begin{figure}[t!]
  \centering
  \includegraphics[width=1.0\linewidth]{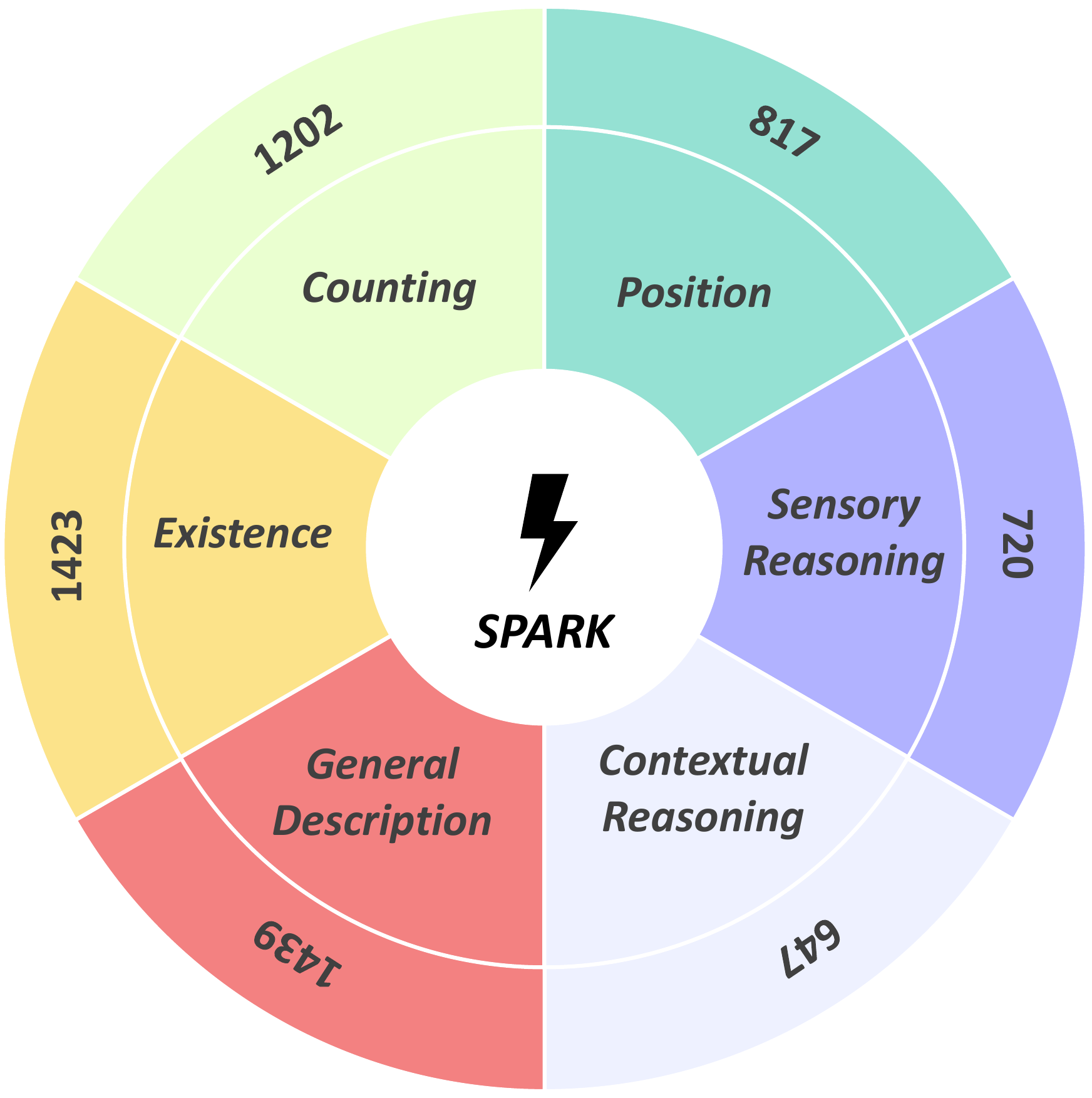}
  \caption{Distribution of data sources of the SPARK benchmark. In SPARK, we demonstrate six core multi-vision sensory tasks in the inner ring, and the outer ring displays the number of samples for each specific task. }
  \vspace{-0.4cm}
  \label{fig:1}
\end{figure}
%%%%%%%%%%%%%%%%%%%%%%%%%%%%%%%%%%%%%%%%%%%%%%%%%%%%%%%%%%%%

\section{Evaluation and Instruction Design}
There are multiple formats available for evaluating the multi-sensor perception and reasoning capabilities of LVLM, each with distinct advantages and limitations. Free-form questions~\cite{yarom2024you} offer flexibility and ease of creation but demand labor-intensive human assessment and present challenges in maintaining consistent scoring. Similarity-based assessment are less resource-intensive but can be significantly affected by biases present in the similarity metrics. Yes-or-No questions~\cite{fu2024mmecomprehensiveevaluationbenchmark} are straightforward and easier to assess, but they may oversimplify the evaluation, failing to capture the full extent of LVLM's comprehension of multi-vision reasoning ability.

First of all, to enable quantitative performance metrics for multi-vision perception, the instruction design aims to elicit ``yes" or ``no" responses from the model. This binary response format simplifies the evaluation process, allowing for clear, objective performance measurement. As a result, each instruction comprises two parts: a brief, targeted question and an explanation corresponding to either ``yes" or ``no." This structure ensures that the LVLM's comprehension can be precisely assessed. For every test image, two instructions are manually crafted, each posing a different question to the model. These questions are designed to test different aspects of the image's content and context. The rationale behind this approach is to ensure that the model's answers are not based on chance. When the LVLMs correctly answer both questions, it demonstrates an understanding of the image and its related information, rather than merely guessing.

\begin{table*}[t!]
\centering
	\renewcommand{\arraystretch}{1.1}
	\renewcommand{\tabcolsep}{2mm}
\resizebox{1.0\linewidth}{!}{
\begin{tabular}{llcccccccc}
\Xhline{5\arrayrulewidth}
Models                          & Vision Sensors & Existence & Count & Position & \begin{tabular}[c]{@{}c@{}}General\\ Description\end{tabular} & \cellcolor[HTML]{EFEFEF}\begin{tabular}[c]{@{}c@{}}Multi-vision \\ Perception\end{tabular} & \begin{tabular}[c]{@{}c@{}}Contextual \\ Reasoning\end{tabular} & \begin{tabular}[c]{@{}c@{}}Sensory \\ Reasoning\end{tabular} & \cellcolor[HTML]{EFEFEF}\begin{tabular}[c]{@{}c@{}}Multi-vision \\ Reasoning\end{tabular} \\ \Xhline{5\arrayrulewidth}
\multicolumn{10}{c}{Open Source Large-scale Vision-Language Models}                                                                                                                                                                                                                                                                                                                                                                                                                                                 \\\Xhline{3\arrayrulewidth}

& RGB                                                      & 93.9      & 68.5  & 62.6     & 97.9                                                          & \cellcolor[HTML]{EFEFEF}80.7                                                               & 95.1                                                            & 97.2                                                         & \cellcolor[HTML]{EFEFEF}96.1                                                              \\

\multirow{-2}{*}{Qwen-VL-Chat}       & Thermal                                                  & 86.1      & 66.9  & 59.3     & 95.3                                                          & \cellcolor[HTML]{EFEFEF}76.9                                                               & 90.3                                                            & 83.5                                                         & \cellcolor[HTML]{EFEFEF}86.9                                                              \\
\multirow{-2}{*}{\cite{bai2023qwenvlversatilevisionlanguagemodel}}                                & Depth                                                    & 76.6      & 59.6  & 53.3     & 84.9                                                          & \cellcolor[HTML]{EFEFEF}68.6                                                               & 78.1                                                            & 68.4                                                         & \cellcolor[HTML]{EFEFEF}73.3                                                              \\
              & XR                                                       & 68.0      & 71.3  & 55.1     & 74.1                                                          & \cellcolor[HTML]{EFEFEF}67.1                                                               & 81.8                                                            & 74.7                                                         & \cellcolor[HTML]{EFEFEF}78.3                                                              \\ 
               \Xhline{3\arrayrulewidth}

& RGB                                                      & 94.2      & 75.5  & 59.8     & 96.9                                                          & \cellcolor[HTML]{EFEFEF}81.6                                                               & 88.7                                                            & 94.8                                                         & \cellcolor[HTML]{EFEFEF}91.8                                                             \\

\multirow{-2}{*}{LLAVA-v1.5-7B}         & Thermal                                                 & 93.3      & 76.1  & 62.4     & 95.1                                                          & \cellcolor[HTML]{EFEFEF}81.7                                                               & 85.5                                                            & 51.0                                                         & \cellcolor[HTML]{EFEFEF}68.2                                                             \\
\multirow{-2}{*}{\cite{liu2023improvedllava}}                                & Depth                                                  & 87.1      & 70.7  & 53.3     & 93.7                                                          & \cellcolor[HTML]{EFEFEF}76.2                                                               & 87.4                                                            & 73.8                                                         & \cellcolor[HTML]{EFEFEF}80.6                                                             \\
              & XR                                                       & 74.2      & 57.4  & 67.4     & 72.3                                                          & \cellcolor[HTML]{EFEFEF}67.8                                                               & 62.1                                                            & 50.7              

              & \cellcolor[HTML]{EFEFEF}56.4                                                              \\ \Xhline{3\arrayrulewidth}

                                & RGB                                                      & 96.5      & 73.4  & 61.4     & 97.2                                                          & \cellcolor[HTML]{EFEFEF}82.1                                                               & 98.0                                                            & 97.2                                                         & \cellcolor[HTML]{EFEFEF}97.6                                                              \\
\multirow{-2}{*}{CogVLM-Chat}        & Thermal                                                 & 94.9      & 76.1  & 64.6     & 96.2                                                          & \cellcolor[HTML]{EFEFEF}82.9                                                               & 96.2                                                            & 59.0                                                         & \cellcolor[HTML]{EFEFEF}77.6                                                              \\
\multirow{-2}{*}{\cite{wang2023cogvlm}}                                & Depth                                                   & 94.9      & 76.1  & 64.5     & 96.5                                                          & \cellcolor[HTML]{EFEFEF}83.0                                                               & 90.1                                                            & 71.7                                                         & \cellcolor[HTML]{EFEFEF}80.9                                                              \\
              & XR                                                       & 86.1      & 72.8  & 61.6     & 79.4                                                          & \cellcolor[HTML]{EFEFEF}74.9                                                               & 90.9                                                            & 84.0                                                         & \cellcolor[HTML]{EFEFEF}87.5                                                              \\ \Xhline{3\arrayrulewidth}
                                & RGB                                                      & 97.2      & 78.5  & 72.2     & 97.9                                                          & \cellcolor[HTML]{EFEFEF}86.4                                                               & 98.0                                                            & 99.5                                                         & \cellcolor[HTML]{EFEFEF}98.8                                                              \\
\multirow{-2}{*}{Meteor-7B}        & Thermal                                                 & 93.5      & 68.9  & 71.7     & 95.3                                                          & \cellcolor[HTML]{EFEFEF}82.3                                                               & 90.9                                                            & 62.0                                                         & \cellcolor[HTML]{EFEFEF}76.4                                                              \\
\multirow{-2}{*}{\cite{lee2024meteormambabasedtraversalrationale}}                                & Depth                                               & 83.5      & 65.9  & 62.2     & 91.6                                                          & \cellcolor[HTML]{EFEFEF}75.8                                                               & 89.5                                                            & 77.3                                                         & \cellcolor[HTML]{EFEFEF}83.4                                                              \\
\multirow{-2}{*}{}              & XR                                                       & 79.5      & 70.6  & 63.8     & 76.6                                                          & \cellcolor[HTML]{EFEFEF}72.6                                                               & 86.4                                                            & 84.0                                                         & \cellcolor[HTML]{EFEFEF}85.2                                                              \\ \Xhline{3\arrayrulewidth} 
                                & RGB                                                      & 96.9      & 81.2  & 69.3     & 96.5                                                          & \cellcolor[HTML]{EFEFEF}85.9                                                               & 98.0                                                            & 99.5                                                         & \cellcolor[HTML]{EFEFEF}98.8                                                              \\
\multirow{-2}{*}{TroL-7B}          & Thermal                                                  & 93.9      & 72.8  & 68.1     & 92.8                                                          & \cellcolor[HTML]{EFEFEF}81.9                                                               & 94.1                                                            & 65.5                                                         & \cellcolor[HTML]{EFEFEF}79.8                                                              \\
\multirow{-2}{*}{\cite{lee2024troltraversallayerslarge}}                                 & Depth                                                   & 83.3      & 67.7  & 67.3     & 90.7                                                          & \cellcolor[HTML]{EFEFEF}77.2                                                               & 84.8                                                            & 73.8                                                         & \cellcolor[HTML]{EFEFEF}79.3                                                              \\
\multirow{-2}{*}{}              & XR                                                       & 82.8      & 69.1  & 71.0     & 78.7                                                          & \cellcolor[HTML]{EFEFEF}75.4                                                               & 83.3                                                            & 84.0                                                         & \cellcolor[HTML]{EFEFEF}83.7                                                              \\  \Xhline{3\arrayrulewidth}
                                & RGB                                                      & 96.5      & 76.9  & 69.3     & 98.6                                                          & \cellcolor[HTML]{EFEFEF}85.3                                                               & 98.6                                                            & 99.5                                                         & \cellcolor[HTML]{EFEFEF}99.1                                                              \\
\multirow{-2}{*}{IXC2.5-VL-7B}        & Thermal                                                 & 93.0      & 70.6  & 66.8     & 95.5                                                          & \cellcolor[HTML]{EFEFEF}81.5                                                               & 92.5                                                            & 60.0                                                         & \cellcolor[HTML]{EFEFEF}76.2                                                              \\
\multirow{-2}{*}{\cite{zhang2024internlmxcomposer25versatilelargevision}}                                 & Depth                                                    & 86.1      & 59.9  & 59.4     & 93.3                                                          & \cellcolor[HTML]{EFEFEF}74.7                                                               & 90.6                                                            & 74.3                                                         & \cellcolor[HTML]{EFEFEF}82.4                                                              \\
\multirow{-2}{*}{}              & XR                                                       & 86.1      & 73.5  & 63.8     & 76.6                                                          & \cellcolor[HTML]{EFEFEF}75.0                                                               & 89.4                                                            & 88.0                                                         & \cellcolor[HTML]{EFEFEF}88.7                                                              \\ \Xhline{3\arrayrulewidth}

                                & RGB                                                      & 97.2      & 78.3  & 72.4     & 97.9                                                          & \cellcolor[HTML]{EFEFEF}86.5                                                               & 97.6                                                            & 99.1                                                         & \cellcolor[HTML]{EFEFEF}98.3                                                              \\
\multirow{-2}{*}{InternVL2-8B}     & Thermal                                                  & 90.5      & 75.8  & 61.1     & 93.7                                                          & \cellcolor[HTML]{EFEFEF}80.3                                                               & 94.6                                                            & 61.5                                                         & \cellcolor[HTML]{EFEFEF}78.1                                                              \\
\multirow{-2}{*}{\cite{internvl2blog}}                                & Depth                                                    & 83.0      & 60.2  & 60.3     & 91.4                                                          & \cellcolor[HTML]{EFEFEF}73.7                                                               & 86.9                                                            & 79.9                                                         & \cellcolor[HTML]{EFEFEF}83.5                                                              \\
\multirow{-2}{*}{}              & XR                                                       & 92.7      & 77.9  & 71.7     & 84.9                                                          & \cellcolor[HTML]{EFEFEF}81.8                                                               & 89.4                                                            & 82.7                                                         & \cellcolor[HTML]{EFEFEF}86.0                                                              \\ \Xhline{3\arrayrulewidth}

\multicolumn{10}{c}{Closed Source Large-scale Vision-Language Models}                                                                                                                                                                                                                                                                                                                                                                                                                                               \\ \Xhline{3\arrayrulewidth}

 & RGB                                                      & 94.6      & 79.6  & 65.2     & 95.3                                                          & \cellcolor[HTML]{EFEFEF}83.7                                                               & 97.6                                                            & 98.6                                                         & \cellcolor[HTML]{EFEFEF}98.1                                                              \\
\multirow{-2}{*}{Gemini 1.5 Pro} & Thermal                                                  & 91.4      & 73.6  & 68.8     & 93.9                                                          & \cellcolor[HTML]{EFEFEF}81.9                                                               & 90.3                                                            & 93.0                                                         & \cellcolor[HTML]{EFEFEF}91.7                                                              \\
\multirow{-2}{*}{\cite{reid2024gemini1_5}}                                & Depth                                                  & 87.8      & 73.7  & 62.6     & 94.2                                                          & \cellcolor[HTML]{EFEFEF}79.6                                                               & 78.0                                                            & 88.4                                                         & \cellcolor[HTML]{EFEFEF}83.2                                                              \\
\multirow{-2}{*}{}              & XR                                                       & 89.9      & 81.6  & 63.0     & 82.0                                                          & \cellcolor[HTML]{EFEFEF}79.2                                                               & 92.4                                                            & 88.0                                                         & \cellcolor[HTML]{EFEFEF}90.2                                                              \\ \Xhline{3\arrayrulewidth}

                                & RGB                                                      & 95.1      & 79.0  & 69.7     & 95.8                                                          & \cellcolor[HTML]{EFEFEF}84.9                                                               & 99.5                                                            & 97.2                                                         & \cellcolor[HTML]{EFEFEF}98.3                                                              \\
\multirow{-2}{*}{Claude 3.5 Sonnet}    & Thermal                                                  & 92.1      & 79.2  & 62.9     & 95.0                                                          & \cellcolor[HTML]{EFEFEF}82.3                                                               & 94.1                                                            & 85.0                                                         & \cellcolor[HTML]{EFEFEF}89.6                                                              \\
\multirow{-2}{*}{\cite{claude3.5sonnet_blog}}                                & Depth                                                    & 72.9      & 67.7  & 55.6     & 84.4                                                          & \cellcolor[HTML]{EFEFEF}70.2                                                               & 86.4                                                            & 75.5                                                         & \cellcolor[HTML]{EFEFEF}80.9                                                              \\
\multirow{-2}{*}{}              & XR                                                       & 83.2      & 76.5  & 74.6     & 83.5                                                          & \cellcolor[HTML]{EFEFEF}79.5                                                               & 93.9                                                            & 82.7                                                         & \cellcolor[HTML]{EFEFEF}88.3                                                              \\ \Xhline{3\arrayrulewidth}

& RGB                                                      & 96.9      & 80.9  & 71.4     & 97.4                                                          & \cellcolor[HTML]{EFEFEF}86.7                                                               & 98.5                                                            & 98.6                                                         & \cellcolor[HTML]{EFEFEF}98.6                                                              \\
\multirow{-2}{*}{GPT-4o}                                & Thermal                                                 & 96.1      & 75.6  & 71.4     & 98.2                                                          & \cellcolor[HTML]{EFEFEF}85.3                                                               & 95.2                                                            & 92.0                                                         & \cellcolor[HTML]{EFEFEF}93.6                                                              \\
\multirow{-2}{*}{\cite{gpt4oblog}}                                & Depth                                                    & 87.6      & 77.3  & 71.0     & 94.4                                                          & \cellcolor[HTML]{EFEFEF}82.6                                                               & 95.8                                                            & 85.8                                                         & \cellcolor[HTML]{EFEFEF}90.8                                                              \\
              & XR                                                       & 91.9      & 83.8  & 65.2     & 85.6                                                          & \cellcolor[HTML]{EFEFEF}81.7                                                               & 95.5                                                            & 82.7                                                         & \cellcolor[HTML]{EFEFEF}89.1                                                              \\ \Xhline{5\arrayrulewidth}
                               
\end{tabular}}
\caption{Evaluation results of different models on SPARK benchmark. Accuracy is the metric. ``Multi-vision Perception" shows the average performance on four dimensions (Existence, Count, Position, and General Description) for evaluating visual perception, and ``Multi-vision Reasoning” shows the average performance on two dimensions (Contextual Reasoning and Sensory Reasoning) for evaluating vision sensory understanding. LVLMs are sorted in ascending order of release date.}
\end{table*}

% \usepackage{multirow}
\begin{table*}[t!]
\centering
	\renewcommand{\arraystretch}{1.1}
	\renewcommand{\tabcolsep}{1mm}
\resizebox{1.0\linewidth}{!}{
\begin{tabular}{lccccccccc}
\Xhline{5\arrayrulewidth}
\multicolumn{1}{l|}{Vision Sensors}                         & \multicolumn{2}{c|}{RGB}                                                                                                                                 & \multicolumn{2}{c|}{Thermal}                                                                                                                             & \multicolumn{2}{c|}{Depth}                                                                                                                               & \multicolumn{2}{c|}{XR}                                                                                                                                  &                                                         \\ \cline{1-9}
\multicolumn{1}{l|}{Models}                                 & \begin{tabular}[c]{@{}c@{}} Multi-Vison \\  Perception\end{tabular} & \multicolumn{1}{c|}{\begin{tabular}[c]{@{}c@{}} Multi-Vision\\  Reasoning\end{tabular}} & \begin{tabular}[c]{@{}c@{}}Multi-Vison\\ Perception\end{tabular} & \multicolumn{1}{c|}{\begin{tabular}[c]{@{}c@{}}Multi-Vision\\ Reasoning\end{tabular}} & \begin{tabular}[c]{@{}c@{}}Multi-Vison\\ Perception\end{tabular} & \multicolumn{1}{c|}{\begin{tabular}[c]{@{}c@{}}Multi-Vision\\ Reasoning\end{tabular}} & \begin{tabular}[c]{@{}c@{}}Multi-Vison\\ Perception\end{tabular} & \multicolumn{1}{c|}{\begin{tabular}[c]{@{}c@{}}Multi-Vision\\ Reasoning\end{tabular}} & \hspace{15pt} \multirow{-2}{*}{ALL} \hspace{15pt}                                    \\ \Xhline{3\arrayrulewidth}
\multicolumn{10}{c}{Open Source Large-scale Vision-Language Models}                                                                                                                                                                                                                                                                                                                                                                                                                                                                                                                                                                                                                                                                                               \\ \Xhline{3\arrayrulewidth}
\rowcolor[HTML]{EFEFEF} 
\multicolumn{1}{l|}{\cellcolor[HTML]{EFEFEF}LLaVA-v1.5-7B}  & \cellcolor[HTML]{EFEFEF}                                         & \multicolumn{1}{c|}{\cellcolor[HTML]{EFEFEF}}                                         & \cellcolor[HTML]{EFEFEF}                                         & \multicolumn{1}{c|}{\cellcolor[HTML]{EFEFEF}}                                         & \cellcolor[HTML]{EFEFEF}                                         & \multicolumn{1}{c|}{\cellcolor[HTML]{EFEFEF}}                                         & \cellcolor[HTML]{EFEFEF}                                         & \multicolumn{1}{c|}{\cellcolor[HTML]{EFEFEF}}                                         & \cellcolor[HTML]{EFEFEF}                                \\
\rowcolor[HTML]{EFEFEF} 
\multicolumn{1}{l|}{\cellcolor[HTML]{EFEFEF}\cite{liu2023improvedllava}}               & \multirow{-2}{*}{\cellcolor[HTML]{EFEFEF}81.6}                   & \multicolumn{1}{c|}{\multirow{-2}{*}{\cellcolor[HTML]{EFEFEF}91.8}}                   & \multirow{-2}{*}{\cellcolor[HTML]{EFEFEF}81.7}                   & \multicolumn{1}{c|}{\multirow{-2}{*}{\cellcolor[HTML]{EFEFEF}68.2}}                   & \multirow{-2}{*}{\cellcolor[HTML]{EFEFEF}76.2}                   & \multicolumn{1}{c|}{\multirow{-2}{*}{\cellcolor[HTML]{EFEFEF}80.6}}                   & \multirow{-2}{*}{\cellcolor[HTML]{EFEFEF}67.8}                   & \multicolumn{1}{c|}{\multirow{-2}{*}{\cellcolor[HTML]{EFEFEF}56.4}}                   & \multirow{-2}{*}{\cellcolor[HTML]{EFEFEF}75.6}          \\
\multicolumn{1}{l|}{Qwen-VL-Chat}                           &                                                                  & \multicolumn{1}{c|}{}                                                                 &                                                                  & \multicolumn{1}{c|}{}                                                                 &                                                                  & \multicolumn{1}{c|}{}                                                                 &                                                                  & \multicolumn{1}{c|}{}                                                                 &                                                         \\
\multicolumn{1}{l|}{\cite{bai2023qwenvlversatilevisionlanguagemodel}}                                       & \multirow{-2}{*}{80.7}                                           & \multicolumn{1}{c|}{\multirow{-2}{*}{96.1}}                                           & \multirow{-2}{*}{76.9}                                           & \multicolumn{1}{c|}{\multirow{-2}{*}{\textbf{\underline{86.9}}}}                                  & \multirow{-2}{*}{68.6}                                           & \multicolumn{1}{c|}{\multirow{-2}{*}{73.3}}                                           & \multirow{-2}{*}{67.1}                                           & \multicolumn{1}{c|}{\multirow{-2}{*}{78.3}}                                           & \multirow{-2}{*}{78.5}                                  \\
\rowcolor[HTML]{EFEFEF} 
\multicolumn{1}{l|}{\cellcolor[HTML]{EFEFEF}Meteor-7B}      & \cellcolor[HTML]{EFEFEF}                                         & \multicolumn{1}{c|}{\cellcolor[HTML]{EFEFEF}}                                         & \cellcolor[HTML]{EFEFEF}                                         & \multicolumn{1}{c|}{\cellcolor[HTML]{EFEFEF}}                                         & \cellcolor[HTML]{EFEFEF}                                         & \multicolumn{1}{c|}{\cellcolor[HTML]{EFEFEF}}                                         & \cellcolor[HTML]{EFEFEF}                                         & \multicolumn{1}{c|}{\cellcolor[HTML]{EFEFEF}}                                         & \cellcolor[HTML]{EFEFEF}                                \\
\rowcolor[HTML]{EFEFEF} 
\multicolumn{1}{l|}{\cellcolor[HTML]{EFEFEF}\cite{lee2024meteormambabasedtraversalrationale}}               & \multirow{-2}{*}{\cellcolor[HTML]{EFEFEF}86.4}                   & \multicolumn{1}{c|}{\multirow{-2}{*}{\cellcolor[HTML]{EFEFEF}98.8}}                   & \multirow{-2}{*}{\cellcolor[HTML]{EFEFEF}82.3}                   & \multicolumn{1}{c|}{\multirow{-2}{*}{\cellcolor[HTML]{EFEFEF}76.4}}                   & \multirow{-2}{*}{\cellcolor[HTML]{EFEFEF}75.8}                   & \multicolumn{1}{c|}{\multirow{-2}{*}{\cellcolor[HTML]{EFEFEF}83.4}}                   & \multirow{-2}{*}{\cellcolor[HTML]{EFEFEF}72.6}                   & \multicolumn{1}{c|}{\multirow{-2}{*}{\cellcolor[HTML]{EFEFEF}85.2}}                   & \multirow{-2}{*}{\cellcolor[HTML]{EFEFEF}82.6}          \\
\multicolumn{1}{l|}{TroL-7B}                                &                                                                  & \multicolumn{1}{c|}{}                                                                 &                                                                  & \multicolumn{1}{c|}{}                                                                 &                                                                  & \multicolumn{1}{c|}{}                                                                 &                                                                  & \multicolumn{1}{c|}{}                                                                 &                                                         \\
\multicolumn{1}{l|}{\cite{lee2024troltraversallayerslarge}}                                       & \multirow{-2}{*}{85.9}                                           & \multicolumn{1}{c|}{\multirow{-2}{*}{98.8}}                                           & \multirow{-2}{*}{81.9}                                           & \multicolumn{1}{c|}{\multirow{-2}{*}{79.8}}                                           & \multirow{-2}{*}{77.2}                                           & \multicolumn{1}{c|}{\multirow{-2}{*}{79.3}}                                           & \multirow{-2}{*}{75.4}                                           & \multicolumn{1}{c|}{\multirow{-2}{*}{83.7}}                                           & \multirow{-2}{*}{82.8}                                  \\
\rowcolor[HTML]{EFEFEF} 
\multicolumn{1}{l|}{\cellcolor[HTML]{EFEFEF}IXC2.5-VL-7B}   & \cellcolor[HTML]{EFEFEF}                                         & \multicolumn{1}{c|}{\cellcolor[HTML]{EFEFEF}}                                         & \cellcolor[HTML]{EFEFEF}                                         & \multicolumn{1}{c|}{\cellcolor[HTML]{EFEFEF}}                                         & \cellcolor[HTML]{EFEFEF}                                         & \multicolumn{1}{c|}{\cellcolor[HTML]{EFEFEF}}                                         & \cellcolor[HTML]{EFEFEF}                                         & \multicolumn{1}{c|}{\cellcolor[HTML]{EFEFEF}}                                         & \cellcolor[HTML]{EFEFEF}                                \\
\rowcolor[HTML]{EFEFEF} 
\multicolumn{1}{l|}{\cellcolor[HTML]{EFEFEF}\cite{zhang2024internlmxcomposer25versatilelargevision}}               & \multirow{-2}{*}{\cellcolor[HTML]{EFEFEF}85.3}                   & \multicolumn{1}{c|}{\multirow{-2}{*}{\cellcolor[HTML]{EFEFEF}\textbf{\underline{99.1}}}}          & \multirow{-2}{*}{\cellcolor[HTML]{EFEFEF}81.5}                   & \multicolumn{1}{c|}{\multirow{-2}{*}{\cellcolor[HTML]{EFEFEF}76.2}}                   & \multirow{-2}{*}{\cellcolor[HTML]{EFEFEF}74.7}                   & \multicolumn{1}{c|}{\multirow{-2}{*}{\cellcolor[HTML]{EFEFEF}82.4}}                   & \multirow{-2}{*}{\cellcolor[HTML]{EFEFEF}75.0}                   & \multicolumn{1}{c|}{\multirow{-2}{*}{\cellcolor[HTML]{EFEFEF}\textbf{\underline{88.7}}}}          & \multirow{-2}{*}{\cellcolor[HTML]{EFEFEF}82.9}          \\
\multicolumn{1}{l|}{CogVLM-Chat}                            &                                                                  & \multicolumn{1}{c|}{}                                                                 &                                                                  & \multicolumn{1}{c|}{}                                                                 &                                                                  & \multicolumn{1}{c|}{}                                                                 &                                                                  & \multicolumn{1}{c|}{}                                                                 &                                                         \\
\multicolumn{1}{l|}{\cite{wang2023cogvlm}}                                       & \multirow{-2}{*}{82.1}                                           & \multicolumn{1}{c|}{\multirow{-2}{*}{97.6}}                                           & \multirow{-2}{*}{\textbf{\underline{82.9}}}                                 & \multicolumn{1}{c|}{\multirow{-2}{*}{77.6}}                                           & \multirow{-2}{*}{\textbf{\underline{83.0}}}                                  & \multicolumn{1}{c|}{\multirow{-2}{*}{80.9}}                                           & \multirow{-2}{*}{74.9}                                           & \multicolumn{1}{c|}{\multirow{-2}{*}{87.5}}                                           & \multirow{-2}{*}{83.3}                                  \\
\rowcolor[HTML]{EFEFEF} 
\multicolumn{1}{l|}{\cellcolor[HTML]{EFEFEF}InternVL2-8B}   & \cellcolor[HTML]{EFEFEF}                                         & \multicolumn{1}{c|}{\cellcolor[HTML]{EFEFEF}}                                         & \cellcolor[HTML]{EFEFEF}                                         & \multicolumn{1}{c|}{\cellcolor[HTML]{EFEFEF}}                                         & \cellcolor[HTML]{EFEFEF}                                         & \multicolumn{1}{c|}{\cellcolor[HTML]{EFEFEF}}                                         & \cellcolor[HTML]{EFEFEF}                                         & \multicolumn{1}{c|}{\cellcolor[HTML]{EFEFEF}}                                         & \cellcolor[HTML]{EFEFEF}                                \\
\rowcolor[HTML]{EFEFEF} 
\multicolumn{1}{l|}{\cellcolor[HTML]{EFEFEF}\cite{internvl2blog}}               & \multirow{-2}{*}{\cellcolor[HTML]{EFEFEF}\textbf{\underline{86.5}}}          & \multicolumn{1}{c|}{\multirow{-2}{*}{\cellcolor[HTML]{EFEFEF}98.3}}                   & \multirow{-2}{*}{\cellcolor[HTML]{EFEFEF}80.3}                   & \multicolumn{1}{c|}{\multirow{-2}{*}{\cellcolor[HTML]{EFEFEF}78.1}}                   & \multirow{-2}{*}{\cellcolor[HTML]{EFEFEF}73.7}                   & \multicolumn{1}{c|}{\multirow{-2}{*}{\cellcolor[HTML]{EFEFEF}\textbf{\underline{83.5}}}}          & \multirow{-2}{*}{\cellcolor[HTML]{EFEFEF}\textbf{\underline{81.8}}}          & \multicolumn{1}{c|}{\multirow{-2}{*}{\cellcolor[HTML]{EFEFEF}86.0}}                   & \multirow{-2}{*}{\cellcolor[HTML]{EFEFEF}\textbf{\underline{83.5}}} \\ \Xhline{3\arrayrulewidth}
\multicolumn{10}{c}{Closed Source Large-scale Vision-Language Models}                                                                                                                                                                                                                                                                                                                                                                                                                                                                                                                                                                                                                                                                                             \\ \Xhline{3\arrayrulewidth}
\multicolumn{1}{l|}{Claude 3.5 Sonnet}                      &                                                                  & \multicolumn{1}{c|}{}                                                                 &                                                                  & \multicolumn{1}{c|}{}                                                                 &                                                                  & \multicolumn{1}{c|}{}                                                                 &                                                                  & \multicolumn{1}{c|}{}                                                                 &                                                         \\
\multicolumn{1}{l|}{\cite{claude3.5sonnet_blog}}                                       & \multirow{-2}{*}{84.9}                                           & \multicolumn{1}{c|}{\multirow{-2}{*}{98.3}}                                           & \multirow{-2}{*}{82.3}                                           & \multicolumn{1}{c|}{\multirow{-2}{*}{89.6}}                                           & \multirow{-2}{*}{70.2}                                           & \multicolumn{1}{c|}{\multirow{-2}{*}{80.9}}                                           & \multirow{-2}{*}{79.5}                                           & \multicolumn{1}{c|}{\multirow{-2}{*}{88.3}}                                           & \multirow{-2}{*}{84.3}                                  \\
\rowcolor[HTML]{EFEFEF} 
\multicolumn{1}{l|}{\cellcolor[HTML]{EFEFEF}Gemini 1.5 Pro} & \cellcolor[HTML]{EFEFEF}                                         & \multicolumn{1}{c|}{\cellcolor[HTML]{EFEFEF}}                                         & \cellcolor[HTML]{EFEFEF}                                         & \multicolumn{1}{c|}{\cellcolor[HTML]{EFEFEF}}                                         & \cellcolor[HTML]{EFEFEF}                                         & \multicolumn{1}{c|}{\cellcolor[HTML]{EFEFEF}}                                         & \cellcolor[HTML]{EFEFEF}                                         & \multicolumn{1}{c|}{\cellcolor[HTML]{EFEFEF}}                                         & \cellcolor[HTML]{EFEFEF}                                \\
\rowcolor[HTML]{EFEFEF} 
\multicolumn{1}{l|}{\cellcolor[HTML]{EFEFEF}\cite{reid2024gemini1_5}}               & \multirow{-2}{*}{\cellcolor[HTML]{EFEFEF}83.7}                   & \multicolumn{1}{c|}{\multirow{-2}{*}{\cellcolor[HTML]{EFEFEF}98.1}}                   & \multirow{-2}{*}{\cellcolor[HTML]{EFEFEF}81.9}                   & \multicolumn{1}{c|}{\multirow{-2}{*}{\cellcolor[HTML]{EFEFEF}91.7}}                   & \multirow{-2}{*}{\cellcolor[HTML]{EFEFEF}79.6}                   & \multicolumn{1}{c|}{\multirow{-2}{*}{\cellcolor[HTML]{EFEFEF}83.2}}                   & \multirow{-2}{*}{\cellcolor[HTML]{EFEFEF}79.2}                   & \multicolumn{1}{c|}{\multirow{-2}{*}{\cellcolor[HTML]{EFEFEF}\textbf{\underline{90.2}}}}          & \multirow{-2}{*}{\cellcolor[HTML]{EFEFEF}85.9}          \\
\multicolumn{1}{l|}{GPT-4o}                                 &                                                                  & \multicolumn{1}{c|}{}                                                                 &                                                                  & \multicolumn{1}{c|}{}                                                                 &                                                                  & \multicolumn{1}{c|}{}                                                                 &                                                                  & \multicolumn{1}{c|}{}                                                                 &                                                         \\
\multicolumn{1}{l|}{\cite{gpt4oblog}}                                       & \multirow{-2}{*}{\textbf{\underline{86.7}}}                                  & \multicolumn{1}{c|}{\multirow{-2}{*}{\textbf{\underline{98.6}}}}                                  & \multirow{-2}{*}{\textbf{\underline{85.3}}}                                  & \multicolumn{1}{c|}{\multirow{-2}{*}{\textbf{\underline{93.6}}}}                                  & \multirow{-2}{*}{\textbf{\underline{82.6}}}                                  & \multicolumn{1}{c|}{\multirow{-2}{*}{\textbf{\underline{90.8}}}}                                  & \multirow{-2}{*}{\textbf{\underline{81.7}}}                                  & \multicolumn{1}{c|}{\multirow{-2}{*}{89.1}}                                           & \multirow{-2}{*}{\textbf{\underline{88.5}}}                         \\ \Xhline{5\arrayrulewidth}
\end{tabular}}
\caption{Leaderboards of 10 advanced leading LVLMs on proposed SPARK benchmark according to different multi-vision sensors. Accuracy is the metric and the best accuracy is denoted in bold and underlined. LVLMs are sorted in ascending order of overall accuracy (ALL).}
\vspace{-0.4cm}
\end{table*}

In addition, we also introduce a multi-vision sensor understanding evaluation design based on multi-choice questions. This format presents questions with a set of predetermined choices, allowing respondents to select the correct options. The multi-choice question format is advantageous for several reasons. First, it enables efficient grading and analysis of responses, as answers can be objectively evaluated against a fixed set of possible responses. Also, the multi-choice question format allows for precise control over the difficulty level of the questions. By varying the validity of each option, we can create questions that test different levels of understanding and comprehension. For example, including more plausible but incorrect options can increase the difficulty, ensuring that only models with a deeper understanding can consistently choose the correct answer. This flexibility in question design makes multi-choice questions a powerful tool for assessing the nuanced capabilities of multi-vision sensor systems. Furthermore, the Yes-or-No format can be seen as a specific case of multi-choice question, where the options are limited to ``(A) Yes" and ``(B) No." This simplification retains the benefits of the multi-choice question format while providing a straightforward way to measure binary decisions. 

Using accuracy as the evaluation metric for both multi-choice questions and Yes-or-No questions ensures consistency in how we assess the model's performance. Accuracy, defined as the proportion of correctly answered questions, provides a clear and intuitive measure of how well the model understands the given inputs. The adoption of the multi-choice question based evaluation design supports the development of a more comprehensive evaluation framework. The incorporation of both simple Yes-or-No questions and more complex multi-choice questions ensures that the evaluation covers both basic and advanced aspects of LVLM's understanding.

\section{Evaluation on Multi-vision Sensor Tasks}
Our instruction dataset was collected according to two multi-vision tasks: multi-vision perception and multi-vision reasoning. As illustrated in Figure 2, first of all, multi-vision perception focuses on the LVLM's ability to accurately interpret and identify objects, scenes, and relationships from various multi-vision inputs. This involves tasks such as object detection, image classification, scene recognition, and relationship detection, where the model must process and understand the content of images from multiple vision sensors. The goal is to ensure that the model can consistently recognize and categorize visual elements across different contexts from different vision sensors.
On the other hand, multi-vision reasoning requires the model to not only perceive but also make inferences based on the multi-vision sensory data. This involves higher-order cognitive tasks such as understanding relationships between objects, prediction of intent of sensor use, and understanding sensor knowledge. For instance, the model might need to infer the cause of an event depicted in an image sequence or predict the purpose of a captured image. Multi-vision reasoning tests the LVLM's capability to integrate multi-vision information with contextual sensory knowledge, making logical deductions that go beyond mere perception.

\subsection{Multi-vision Perception}
Multi-vision perception is the foundational process by which Large Vision-Language Models (LVLMs) analyze images captured by various multi-vision sensors, including RGB, thermal, depth, and X-ray images. This process involves recognizing and interpreting the fundamental elements within each visual input based on cognitive science~\cite{kahneman1992reviewing,broadbent2013perception}.
\begin{itemize}
\item Existence: LVLMs can identify and list common objects present in the image, such as people, vehicles, animals, furniture, and so on. 
\item Count: LVLMs can count the number of identified objects or entities, providing a quantitative understanding of the scene. 
\item Position: LVLMs can determine the spatial arrangement of objects within the image, noting their positions relative to one another. 
\item General Description: LVLMs are also equipped to generate nuanced descriptions of the overall scene depicted in an image. They can articulate what is happening, identify objects, and provide factual information that enhances the understanding of the image itself. 
\end{itemize}
At the perception stage, LVLMs focus on extracting essential information directly from raw image data captured by multi-vision sensors. This foundational perception is critical for all subsequent reasoning tasks, serving as the foundation upon which more complex interpretations are built.

\subsection{Multi-vision Reasoning}
Multi-vision reasoning is where LVLMs truly showcase their advanced capabilities. Beyond simply perceiving images, LVLMs can engage in logical reasoning to derive deeper insights and make informed decisions. This distinguishes the recent LVLMs from traditional computer vision models, which primarily focus on understanding and interacting with the real world. 
\begin{itemize}
\item Contextual reasoning: LVLMs can utilize fundamental knowledge and contextual clues to make judgments about a given scenario. This type of reasoning allows LVLMs to refer to the underlying basis of physical sensor knowledge and ensure that the reasoning process remains consistent with the context provided by the image and the associated information.
\item Sensory reasoning: A more complex reasoning ability requires LVLMs to map 2D image data to the physical meanings associated with different multi-vision sensors. This process not only involves processing the raw data from images but also integrates it with contextual information about the underlying physical sensor knowledge in the real world. By combining fundamental sensor information, LVLMs can derive conclusions that are both accurate and contextually relevant. Sensory reasoning requires a deep understanding of the knowledge underlying the physical meaning of multi-vision sensor data. This goes beyond surface-level image recognition, demanding that LVLMs make sense of the sensor data in a way that reflects real-world physics and usage scenarios.
\end{itemize}

Next, we integrate both visual and textual inputs into GPT-4, guided by meticulously crafted prompts. These prompts are specifically designed to align with various evaluation dimensions, ensuring that the generated questions are both relevant and focused. To further enhance the quality of the benchmark, we introduce an additional filtering step. In the final stages of development, human annotators play a crucial role, selecting the correct answers and categorizing the questions according to their respective evaluation dimensions.

\section{Experiment}
\subsection{Implementation Details}
\subsubsection{Dataset Collection}
We collect six subsets for each multi-sensor vision task type, together with 4k images and 6k unique questions and answers. These instructions are built from five public datasets: MS-COCO \cite{lin2015microsoftcococommonobjects}, M$^3$FD \cite{liu2022targetawaredualadversariallearning}, Dog\&People \cite{thermal-dogs-and-people-x6ejw}, RGB-D scene dataset \cite{cho2021dimlcvlrgbddataset2m}, and UNIFESP X-ray Body Part Classifier Competition dataset~\cite{unifesp-x-ray-body-part-classifier}. The MS-COCO dataset is a commonly used object detection dataset that contains RGB images with fine-grained object bounding boxes, categories, and attribute annotations. We sampled 1.2k images from validation dataset. Furthermore, for thermal sensor datasets, we sampled 1.2k images from two different thermal datasets (M$^3$FD and Dog\&People) in order to collect a thermal dataset covering the widest possible range of diverse situations and objects. Additionally, we sampled 1.2k images from RGB-D scene dataset~\cite{cho2021dimlcvlrgbddataset2m} for depth sensor because it covers a variety of indoor and outdoor scenes. Finally, we sampled 0.4k images from the public X-ray body part dataset for the XR sensor dataset because of the diversity of multiple human body parts. We described the overall distribution of data sources of the SPARK benchmark in Figure 3.

\subsubsection{Large Vision Language Models}
In our evaluation, we selected 10 state-of-the-art (SOTA) Large Vision-Language Models (LVLMs) that represent the leading edge of current research. These models were chosen to provide a comprehensive assessment of the capabilities and performance of both open-source and closed-source LVLMs across a variety of multi-vision sensor tasks on the SPARK benchmark.
\begin{itemize}
\item Open source: CogVLM-Chat~\cite{wang2023cogvlm}, LLAVA-v1.5-7B~\cite{liu2023llava}, InternVL2-8B~\cite{internvl2blog}, TroL-7B~\cite{lee2024troltraversallayerslarge}, Meteor-7B~\cite{lee2024meteormambabasedtraversalrationale}, IXC2.5-VL-7B~\cite{zhang2024internlmxcomposer25versatilelargevision}, Qwen-VL-Chat~\cite{bai2023qwenvlversatilevisionlanguagemodel}
\item Closed source: GPT-4o~\cite{gpt4oblog}, Claude 3.5 Sonnet~\cite{claude3.5sonnet_blog}, Gemini-Pro1.5~\cite{reid2024gemini1_5}
\end{itemize}

\subsection{Experiment Result}
In this section, we conduct a comprehensive evaluation using the proposed SPARK benchmark, a rigorous framework designed to assess the capabilities of Large Vision-Language Models (LVLMs) in two target tasks: Multi-vision Perception and Multi-vision Reasoning. Multi-vision Perception presents the averaged performance on four dimensions for evaluating visual perception. Meanwhile, Multi-vision Reasoning demonstrates the averaged performance on two dimensions for evaluating the LVLMs' ability to understand and reason about multi-vision sensory data. 

As shown in Table 1, the evaluation revealed that performance varies significantly depending on the type of multi-vision sensor used to capture the input images. LVLMs generally perform well in simple Multi-vision perception tasks such as generating general descriptions, but more complex reasoning tasks like Multi-vision Reasoning reveal significant differences in model capabilities. Since they mainly trained with general RGB images, the performance of multi-vision perception and reasoning in RGB sensor is consistently maintained at high levels. However, the performance of LVLMs drops noticeably when dealing with images captured using thermal, depth, and X-ray(XR) sensors. This decline is particularly evident in the Multi-vision Reasoning task, especially in Sensory Reasoning.

Sensory Reasoning requires LVLMs to not only recognize and describe images but also to understand the physical principles underlying the sensor data. For example, interpreting thermal data involves understanding heat signatures, while depth data requires an understanding of the need for spatial geometry beyond simple 2D interpretation. The experiment demonstrates LVLMs' limited proficiency in interpreting and mapping sensor data to its physical meaning.

Table 2 provides a clear comparison of the performance of various LVLMs across different multi-vision sensors and tasks. It highlights the strengths and weaknesses of each model, particularly the advantage that closed-source models have in maintaining high performance across more complex reasoning tasks with diverse vision sensor types. Considering the overall accuracy score (ALL), GPT-4o excels in the proposed SPARK benchmark.

\begin{table}[t!]
\centering
	\renewcommand{\arraystretch}{1.1}
	\renewcommand{\tabcolsep}{1mm}
\resizebox{1.0\linewidth}{!}{
\begin{tabular}{l|l|c|c|c}
\Xhline{4\arrayrulewidth}
Model             & \begin{tabular}[c]{@{}l@{}}Vision\\ Sensor\end{tabular} & \begin{tabular}[c]{@{}c@{}}Sensor Reasoning\\ w/o Sensor Info.\end{tabular} & \begin{tabular}[c]{@{}c@{}}Sensor Reasoning\\ w/ Sensor Info.\end{tabular} & $\Delta$ \\ \Xhline{4\arrayrulewidth}
LLAVA-v1.5-7B     & Thermal                                                 & 51.0                                                                        & 81.0                                                                       & +30.0 \\
\cite{liu2023llava}       & Depth                                                   & 73.8                                                                        & 87.6                                                                       & +13.8 \\
\textit{Open source LVLM}       & XR                                                      & 50.7                                                                        & 54.0                                                                       & +3.3  \\ \hline
TroL-7B           & Thermal                                                 & 65.5                                                                        & 97.0                                                                       & +31.5 \\
\cite{lee2024troltraversallayerslarge}              & Depth                                                   & 73.8                                                                        & 99.1                                                                       & +25.3 \\
\textit{Open source LVLM}       & XR                                                      & 84.0                                                                        & 84.0                                                                       & -     \\ \hline
InternVL2-8B      & Thermal                                                 & 61.5                                                                        & 85.5                                                                       & +24.0 \\
\cite{internvl2blog}            & Depth                                                   & 79.9                                                                        & 99.6                                                                       & +19.7 \\
\textit{Open source LVLM}       & XR                                                      & 82.7                                                                        & 85.3                                                                       & +2.6  \\ \hline
Claude 3.5 Sonnet & Thermal                                                 & 85.0                                                                        & 96.5                                                                       & +11.5 \\
\cite{claude3.5sonnet_blog}            & Depth                                                   & 75.5                                                                        & 99.6                                                                       & +24.1 \\
\textit{Closed source LVLM}     & XR                                                      & 82.7                                                                        & 82.7                                                                       & -     \\ \hline
GPT-4o            & Thermal                                                 & 92.0                                                                        & 94.0                                                                       & +2.0  \\
\cite{gpt4oblog}              & Depth                                                   & 85.8                                                                        & 99.6                                                                       & +13.8 \\
\textit{Closed source LVLM}     & XR                                                      & 82.7                                                                        & 84.0                                                                       & +1.3  \\ \Xhline{4\arrayrulewidth}\end{tabular}}
\caption{Ablation study on sensor reasoning performance change whether the sensor information is given. We choose three LVLMs from open source and two from closed source.}
\vspace{-0.4cm}
\end{table}

\subsection{Ablation study}
In the previous section, we observed that LVLMs frequently struggle to accurately infer the purpose or context of an image when the data is sourced from multi-vision sensors other than RGB. However, as demonstrated in Figure 1, even when the input image lacks explicit information about the sensor type, LVLMs can still identify the sensor correctly. This suggests that while LVLMs have already acquired sensor-related knowledge through textual data, they face challenges in mapping fundamental knowledge to real-world scenarios. 

Thus, in Table 3, we conducted an ablation experiment on data-centric enhancement by adding sensor information as a text prompt at the beginning of the question (``This is a \{Thermal, Depth, X-Ray\} image.") and measured the sensory reasoning performance change. The experiment demonstrated that sensor information can enhance the reasoning capabilities of LVLMs, particularly for thermal and depth images, while XR data showed the least impact. This implies that LVLM models, including GPT-4o, are not fully utilizing the knowledge they already possess to understand multi-vision sensory data.

\section{Conclusion}
In this study, we focus on evaluating the ability of Large Vision-Language Models (LVLMs) to understand and process multi-vision sensory inputs. As LVLMs are increasingly deployed in real-world applications, their ability to accurately interpret and reason about data from diverse vision sensors has become crucial. To address this, we propose an evaluation benchmark called SPARK, which generates instruction tuning samples aimed at specific physical sensor understanding in various question-and-answer formats. Through extensive experiments, we assess the performance of understanding sensory knowledge in the latest state-of-the-art LVLMs handling multi-vision input. We believe this approach, integrating a sensory knowledge annotated evaluation benchmark paves the way for promising future applications of LVLMs.

\bibliography{aaai25}

\end{document}